\documentclass[10pt]{article} 

\usepackage[accepted]{rlj} 

\usepackage{amssymb}            
\usepackage{mathtools}          
\usepackage{mathrsfs}           
\usepackage{graphicx}           
\usepackage{subcaption}         
\usepackage[space]{grffile}     
\usepackage{url}                
\usepackage{lipsum}             

\usepackage{kotex}
\usepackage{amsmath}
\usepackage{multirow}
\usepackage{titletoc}
\usepackage{amssymb}
\usepackage[table, dvipsnames]{xcolor}
\usepackage{goodbad}

\title{V-Simba: Unleashing the Architectural Potential \\ of RL in Visual Continuous Control}

\setrunningtitle{Unleashing the Architectural Potential of RL in Visual Continuous Control}

\author{%
    Donghu Kim\textsuperscript{1},\;%
    Youngdo Lee\textsuperscript{1},\;%
    Hojoon Lee\textsuperscript{1},\;%
    Johan Obando-Ceron\textsuperscript{2,3},\\%
    Byungkun Lee\textsuperscript{1},\;%
    Aaron Courville\textsuperscript{2,3,5},\;%
    Pablo Samuel Castro\textsuperscript{2,3,4},\\%
    Jaegul Choo\textsuperscript{1},\;%
    Clare Lyle\textsuperscript{4}%
}

\emails{quagmire@kaist.ac.kr}

\affiliations{
$^{1}$\textbf{KAIST}\;\;\;
$^{2}$\textbf{Mila – Qu\'{e}bec AI Institute}\;\;\;
$^{3}$\textbf{Universit\'{e} de Montr\'{e}al}\\
$^{4}$\textbf{Google DeepMind}\;\;\;
$^{5}$\textbf{Canada CIFAR AI Chair}
\par 
}

\contribution{
We identify and analyze severe training instabilities such as sharp loss landscapes, dormant units, and feature collapse that are inherent in standard convolutional architectures widely adopted in visual reinforcement learning.
}{
While previous visual RL research has primarily focused on algorithmic innovations, the underlying neural architectures have largely remained simple, often following the standard DrQ-v2 baseline. Our analysis reveals the vulnerabilities and capacity loss associated with these heavily utilized architectures.
}

\contribution{
We propose V-Simba, a novel visual RL architecture integrating normalization layers, weight regularization, and a distributional critic to stabilize training, alongside large-stride and pointwise convolutions to maintain computational efficiency.
}{
Recent state-based models have improved learning by constraining parameters and feature growth, but translating this to pixel-based environments is hindered by the severe computational bottlenecks of high-dimensional observations. V-Simba addresses this gap by combining targeted normalization to stabilize optimization with spatial downsampling to remain lightweight and computationally feasible for visual tasks.
}

\contribution{
We experimentally demonstrate that V-Simba matches or outperforms state-of-the-art visual RL methods across 29 tasks in the DeepMind Control Suite, Adroit, and Meta-World benchmarks.
}{
We evaluate V-Simba against strong algorithmic baselines including DrQ-v2, MR.Q, TD-MPC2, DrM, and TACO, showing that robust architectural design alone can achieve superior sample and compute efficiency without introducing complex algorithmic add-ons.
}

\keywords{Deep RL, Visual RL, Neural Network Design, Plasticity, Normalization} 

\summary{
Improving sample efficiency remains a core challenge in reinforcement learning (RL), especially in real-world settings like robotics, where data collection is costly. 
This challenge is pronounced in visual RL, where high-dimensional inputs often obscure learning signals. 
While prior work in visual RL has focused on algorithmic solutions, such as better dynamics models or exploration strategies, recent advances in state-based RL show that architectural design alone can lead to significant gains in sample efficiency. 
This raises an important question: \textit{Can these architectural principles transfer to visual RL?} In response, we introduce \textbf{V-Simba}, a simple yet effective visual RL architecture inspired by the Simba architecture from state-based RL. 
Built on top of Soft Actor-Critic (SAC) with data augmentation, V-Simba modifies the architecture by adding normalization layers to stabilize training and using pointwise convolutions to reduce computation.
Despite its simplicity, V-Simba matches or outperforms the state-of-the-art methods across the DMC, Adroit, and Meta-World benchmarks, while being more computationally efficient than DrQ-v2.
}

\begin{document}

\makeCover  
\maketitle  

\begin{abstract}
Improving sample efficiency remains a core challenge in reinforcement learning (RL), especially in real-world settings like robotics, where data collection is costly. 
This challenge is pronounced in visual RL, where high-dimensional inputs often obscure learning signals. 
While prior work in visual RL has focused on algorithmic solutions, such as better dynamics models or exploration strategies, recent advances in state-based RL show that architectural design alone can lead to significant gains in sample efficiency. 
This raises an important question: \textit{Can these architectural principles transfer to visual RL?} In response, we introduce \textbf{V-Simba}, a simple yet effective visual RL architecture inspired by the Simba architecture from state-based RL. 
Built on top of Soft Actor-Critic (SAC) with data augmentation, V-Simba modifies the architecture by adding normalization layers to stabilize training and using pointwise convolutions to reduce computation.
Despite its simplicity, V-Simba matches or outperforms the state-of-the-art methods across the DMC, Adroit, and Meta-World benchmarks, while being more computationally efficient than DrQ-v2.
\textbf{We make our code publicly available at} \url{https://github.com/DAVIAN-Robotics/V-Simba}.
\end{abstract}

\vspace{-1mm} 
\begin{figure}[ht!]
\begin{center}
\includegraphics[width=1.0\textwidth]{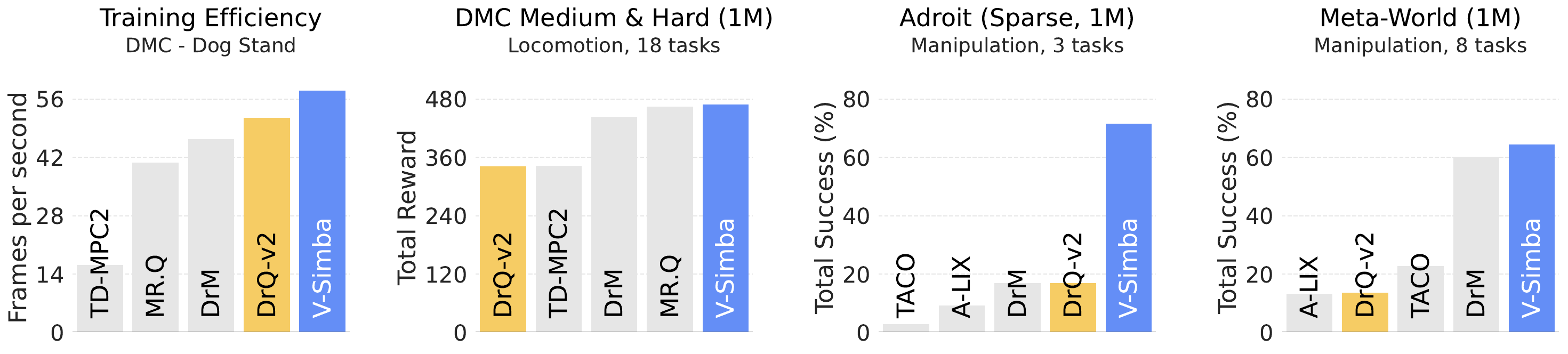}
\caption{\textbf{Benchmark Summary.} We evaluate the effectiveness of V-Simba across 29 visual continuous control tasks spanning multiple domains, with a \textit{single} set of hyperparameters. By incorporating V-Simba into Soft Actor-Critic with data augmentation, it matches or outperforms state-of-the-art visual RL methods, demonstrating better sample and compute efficiency.}
\label{fig:main_performance}
\end{center}
\vspace{-2mm}
\end{figure}

\section{Introduction}
\label{sec:submission}

Deep reinforcement learning (RL) has long been a prominent approach for solving continuous control tasks. However, RL typically relies on an extensive amount of trial-and-error within the environment, which is often expensive in terms of time, compute, and real-world constraints. This issue is exacerbated in visual RL, where agents must learn from high-dimensional, noisy, and often partially observable image inputs. Consequently, improving sample efficiency (i.e., learning effectively from limited interaction data) has become a key research topic in visual RL.

To improve sample efficiency, recent work has largely concentrated on algorithmic innovations, including enhanced representation learning~\citep{yarats2021sacae,echchahed2025survey,obando2026simplicial}, latent dynamics modeling~\citep{fujimoto2025mrq, zheng2023taco}, world models~\citep{hansen2023tdmpcv2, hafner2023dreamerv3}, and improved exploration strategies~\citep{burda2018rnd, xu2023drm}. Yet, despite these advances, the underlying neural architectures have remained relatively simple. 
A prominent example is DrQ-v2~\citep{yarats2021drqv2}, which combines the DDPG algorithm~\citep{lillicrap2015ddpg} with data augmentation~\citep{laskin2020rad}. Its architecture consists of a shallow convolutional encoder, followed by a large fully connected layer and a single layer normalization layer~\citep{lei2016layer_norm} in-between.
Due to its simplicity and strong empirical performance, DrQ-v2 has become the de facto standard in visual RL, and many state-of-the-art methods~\citep{xu2023drm, zheng2023taco, cetin2022alix, sukhija2024maxinforl} adopt DrQ-v2's architecture with minimal modifications.

\begin{figure}[t]
\begin{center}
\includegraphics[width=1.0\textwidth]{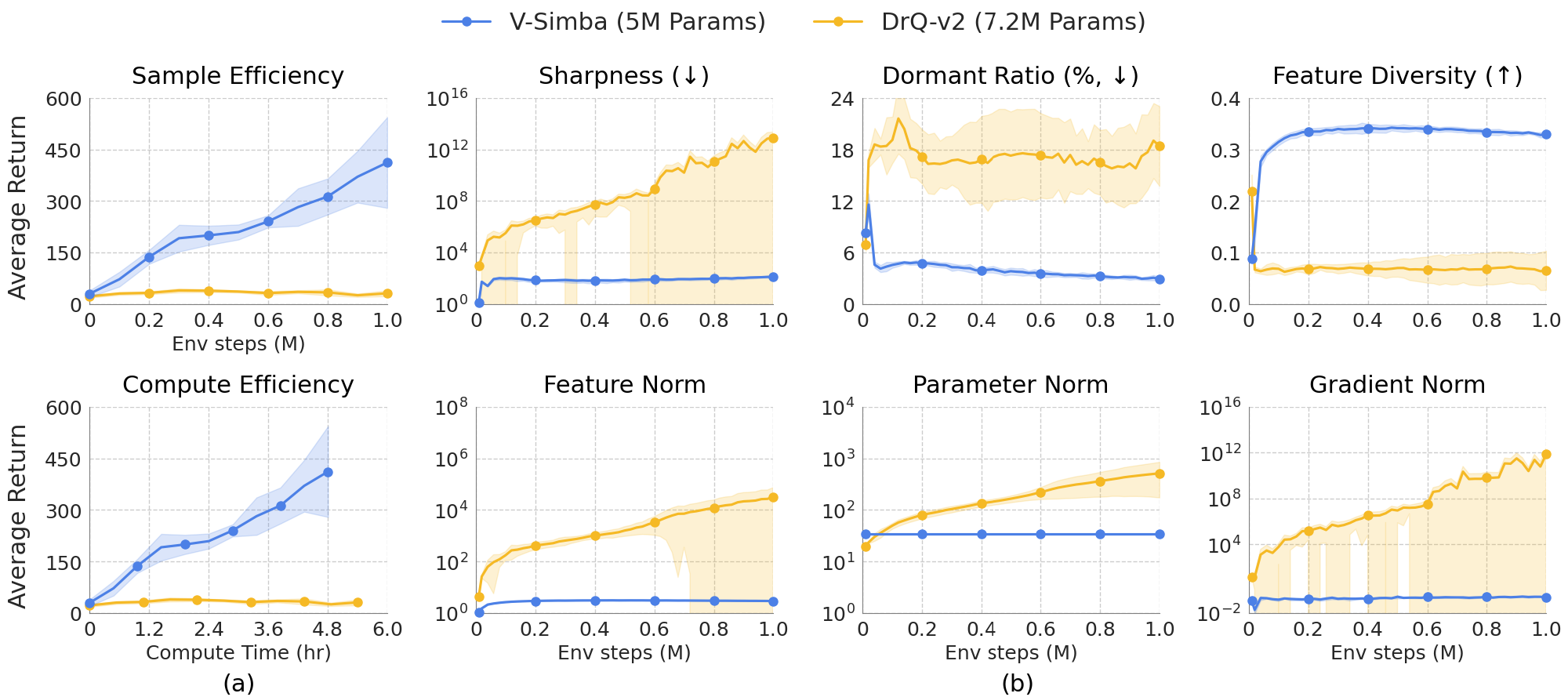}
\vspace{-2mm}
\caption{\textbf{DrQ-v2 vs. V-Simba.} Comparison of the DrQ-v2 architecture and our V-Simba in the \texttt{Dog Stand} environment. Both are evaluated on Soft Actor-Critic (SAC) with results averaged over 5 seeds. \textbf{(a)} V-Simba is substantially more efficient than DrQ-v2 in terms of both sample and compute. \textbf{(b)} Unlike DrQ-v2, V-Simba has stable learning dynamics, indicated by a smooth loss landscape, a low dormant ratio, high feature diversity, and well-controlled feature, parameter, and gradient norms. Detailed explanations of each metric are provided in Appendix~\ref{appendix:metrics}.
}
\label{fig:metric_analysis}
\end{center}
\vspace{-5mm} 
\end{figure}

However, our analysis reveals that this commonly adopted architecture suffers from severe training instabilities. As shown in the top row of Figure~\ref{fig:metric_analysis}(b), DrQ-v2 exhibits sharp loss landscapes that correlate with poor generalization~\citep{foret2020sharpness, lee2024plastic}, a high fraction of dormant units indicating plasticity loss~\citep{sokar2023redo,liu2026measure}, and low feature diversity indicating feature collapse\footnote{We use these metrics as diagnostic probes that surface the failure modes our architecture is designed to address, rather than as direct predictors of final return; we discuss their scope and limitations in Appendix~\ref{appendix:metric_analysis}.}~\citep{woo2023convnextv2}.

In contrast, recent advances in state-based RL~\citep{lee2024simba, bhatt2024crossq, lee2025simbav2, palenicek2025scaling,castanyer2025stable} demonstrate that carefully designed architectures can effectively mitigate these instabilities. Notably, the Simba series of architectures~\citep{lee2024simba, lee2025simbav2} introduce principled architectural guidelines that stabilize training by constraining the growth of features, weights, and gradients through targeted normalization and regularization.

While effective in state-based tasks, the Simba architecture lacks a suitable inductive bias for visual data. Convolutional architectures in RL face distinct challenges, such as severe plasticity loss and capacity degradation driven by unconstrained parameter growth across spatial hierarchies~\citep{lyle2024disentangling}. Furthermore, directly applying extensive normalization to high-resolution convolutional feature maps introduces prohibitive computational bottlenecks.

In response, we propose \textbf{V-Simba}, an architecture specifically tailored to overcome these visual RL challenges. Built on top of Soft Actor-Critic (SAC)~\citep{haarnoja2018sac}, V-Simba incorporates three core architectural components: (1) layer normalization (LN) to control feature norms, (2) $\ell_2$ weight regularization to limit parameter growth, and (3) a distributional critic with reward normalization to stabilize gradients. Crucially, to make these stabilizing mechanisms computationally viable for high-dimensional image inputs, V-Simba employs aggressive early spatial reduction, alongside extensive use of lightweight pointwise convolutions~\citep{hua2018pointwise} and max-pooling operations~\citep{krizhevsky2012imagenet}.

We evaluate V-Simba on three standard benchmarks: DMControl ~\citep{tassa2018dmc}, Adroit~\citep{rajeswaran2017adroit}, and Meta-World~\citep{yu2020metaworld} using a single set of hyperparameters across all tasks. Despite its simplicity, V-Simba significantly outperforms DrQ-v2, while reducing both model size (7.2M → 5.0M) and training time (5.4 → 4.8 hours for 1M DMControl steps). Most notably, V-Simba outperforms highly complex algorithmic methods like DrM~\citep{xu2023drm} across all benchmarks, as well as MR.Q~\citep{fujimoto2025mrq}, TD-MPC2~\citep{hansen2023tdmpcv2} and TACO~\citep{zheng2023taco}.

V-Simba is intended to offer a strong, stable, and efficient architectural foundation for advancing visual continuous control. We hope our work highlights the untapped potential of principled architecture design within the visual RL community.

\section{Related Work}

Learning solely from high-dimensional visual observations poses significant challenges in RL. Due to the partially observable nature of the observation space (Section \ref{preliminary:visualRL}), visual RL agents suffer from poor sample efficiency and large generalization gaps compared to their state-based counterparts~\citep{ma2022visualrlaug}.

\textbf{Algorithmic approaches for visual RL} have primarily focused on: (1) representation learning via auxiliary tasks predicting future latent states, either as auxiliary losses in model-free methods~\citep{stooke2021atc, zheng2023taco, schwarzer2020spr, schwarzer2021sgi, kim2022self, lee2020stochastic, van2016stable, yu2022mask, gelada2019deepmdp, seo2022reinforcement, ni2024bridging, yu2021playvirtual, fujimoto2021deep, mcinroe2021learning, fujimoto2025mrq,
echchahed2025survey,
obando2026simplicial,
pasand2026stable} or separate dynamics models in model-based methods~\citep{hansen2023tdmpcv2, hafner2023dreamerv3, lin2025tdmpcsquared, wu2023daydreamer, ha2018world, finn2016deep, watter2015embed}; (2) data augmentation, especially random shifts~\citep{laskin2020rad, kostrikov2020drq, yarats2021drqv2}, for better efficiency and generalization; and (3) exploration methods, including planning~\citep{sekar2020planning, wang2023coplanner}, curiosity-driven~\citep{pathak2017curiosity, burda2018exploration, guo2022byolexplore}, and information maximization~\citep{sukhija2024maxinforl}. These algorithmic innovations have driven rapid progress in visual RL, leading to continuous improvements in sample efficiency.

\textbf{Architectural design for visual RL} has received comparatively little attention compared to algorithmic innovations. The field has largely maintained shallow convolutional neural network (CNN) architectures similar to the one established by DQN~\citep{mnih2015dqn} over a decade ago. While some works have incorporated architectural elements from computer vision—such as ResNet-like architectures in Impala~\citep{espeholt2018impala}, BBF~\citep{schwarzer2023bbf}, and EfficientZero~\citep{ye2021mastering}, or transformers in DTQN~\citep{esslinger2022deep}—these modifications were often introduced alongside complex algorithmic methods. This entanglement has obscured the true contribution of architectural design to performance improvements. Recent studies have identified the benefits of normalization techniques~\citep{lyle2023understanding_plasticity, ball2023efficient, lyle2024normalization}, but were generally applied to conventional CNN encoders with minimal architectural modifications.

Studies that do revisit visual encoders, such as global average pooling~\citep{trumpp2025impoola, sokar2025mindthegap}, Hadamard max-pooling encoders~\citep{kooi2025hadamax}, or Mixture-of-Experts~\citep{obando2024mixtures, willi2024mixture,sokar2024don}, have largely targeted discrete-action, Atari-style domains. This mirrors a broader resurgence of architectural and representational design across RL, including high-capacity categorical value functions~\citep{nauman2025brc}, dynamic network expansion for scaling continuous control~\citep{kang2025fog,liu2025neuroplastic}, constraining initial representations to stabilize temporal-difference learning~\citep{lyu2026constrained}, and reinitialization schemes that balance the stability-plasticity tradeoff~\citep{han2026fire}. In visual continuous control, however, architectural innovation remains scarce~\citep{espeholt2018impala, huang2024mentor}.

\section{Preliminary}


\subsection{Visual Reinforcement Learning}
\label{preliminary:visualRL}

Reinforcement learning (RL) is typically formulated as a Markov Decision Process (MDP)~\citep{bellman1957markovian}, defined by the tuple $(\mathcal{S}, \mathcal{A}, P, R, \gamma)$ of state space $\mathcal{S}$, action space $\mathcal{A}$, transition function $P: \mathcal{S} \times \mathcal{A} \to \mathcal{P}(\mathcal{S})$, reward function $R: \mathcal{S} \times \mathcal{A} \to \mathbb{R}$, and discount factor $\gamma \in [0, 1)$. From an initial state $s_0 \in \mathcal{S}$, the objective is to find an optimal policy $\pi^*: \mathcal{S} \to \mathcal{P} (\mathcal{A})$ that maximizes the expected discounted return $\mathbb{E}_{\pi} [\sum_{t=0}^{\infty} \gamma^t R(s_t, a_t)]$. Visual RL is a subclass of this problem where the agent does not have access to the true state $s \in \mathcal{S}$, but instead receives high-dimensional pixel observations $o \in \mathcal{O}$ of the system. Since these observations may not fully capture the true state, the problem is modeled as a Partially Observable Markov Decision Process (POMDP)~\citep{bellman1957markovian} represented by the tuple $(\mathcal{S}, \mathcal{O}, \mathcal{A}, P, R, \gamma)$, where $\mathcal{O}$ denotes the observation space. 

\subsection{Data-regularized Q-learning}
\label{preliminary:drq}

Data-regularized Q-learning (DrQ-v2)~\citep{yarats2021drqv2} is a model-free RL algorithm that has emerged as a strong baseline in visual RL due to its simplicity, efficiency, and competitive performance. It builds upon the Deep Deterministic Policy Gradient (DDPG) algorithm \citep{lillicrap2015ddpg} by incorporating two key modifications: (1) extensive use of data augmentation via random shift transformations, and (2) target Q-function stabilization through exponential moving average (EMA) updates. 

At its core, DrQ-v2 improves sample efficiency in off-policy learning by generating augmented views of each observation, thereby increasing data diversity. This augmentation acts as a regularizer, mitigating overfitting to specific visual patterns. Despite its empirical effectiveness, DrQ-v2 employs a notably lightweight architecture: a shallow convolutional encoder followed by an MLP-based prediction head, with a single normalization layer~\citep{lei2016layer_norm} in between.

While DrQ-v2 has become the de facto architecture for many recent visual RL algorithms \citep{xu2023drm, zheng2023taco}, its architectural simplicity leaves room for improvement in stability and representational capacity.

\subsection{Soft Actor-Critic (SAC)}

Soft Actor-Critic (SAC) is a prominent off-policy algorithm for continuous control. It aims to maximize both expected cumulative reward and policy entropy, where $\tau = (o, a, r, o')$ represents a transition tuple. SAC comprises a stochastic policy $\pi_{\theta}(a|o)$, a Q-function $Q_{\phi}(o, a)$, and an entropy coefficient $\alpha$ that balances reward maximization and entropy regularization. The policy network is optimized to maximize the expected return while encouraging exploration through entropy. This objective is formalized as:
\begin{equation} 
\mathcal{L}_{\pi} = \mathbb{E}_{\bar{a} \sim \pi_\theta} \left[ \alpha \log \pi_\theta(\bar{a}|o) - Q_\phi(o, \bar{a}) \right].
\label{eq:policy_objective} \end{equation}

The Q-function $Q_{\phi}(o, a)$ is trained to minimize the Bellman residual: 
\begin{equation} 
\mathcal{L}_Q =(Q_\phi(o, a) - \left( r + \gamma Q_{\bar{\phi}}(o', a')-\alpha\log \pi_\theta(a'|o') \right))^2,
\label{eq:critic_objective} 
\end{equation} 
where $a'\sim \pi_{\theta}(\cdot|o')$, and $Q_{\bar{\phi}}$ represents the target Q-network updated via an exponential moving average of $\phi$.

\section{Method}
\label{section:method}

V-Simba leverages architectural design from state-based RL to stabilize optimization dynamics and improve computational efficiency in visual RL.
Our design follows two core principles: (1) stabilizing optimization (Section~\ref{section:philosophy1}), and (2) maintaining computational efficiency (Section~\ref{section:philosophy2}). The final architecture builds on these principles (Section~\ref{section:vsimba_arch}).

\subsection{Design Philosophy I: Stabilizing Optimization}
\label{section:philosophy1}

As shown in Figure~\ref{fig:metric_analysis}, DrQ-v2 suffers from unstable optimization during training. While layer normalization (LN) \citep{lei2016layer_norm} and residual connections \citep{he2020why_resnet} effectively stabilize supervised learning, visual RL methods often underuse them—DrQ-v2, for instance, employs only a single normalization layer without residuals. We incorporate both components to improve stability.

However, when adding LayerNorm, one must consider its relationship with the gradient. Concretely, LayerNorm introduces scale invariance: for any scalar $c > 0$ and weight matrix $W$,
\begin{equation}
\text{Norm}(cWx) = \text{Norm}(Wx),
\end{equation}
which causes gradients to scale inversely with parameter magnitude:
\begin{equation}
\nabla_W\text{Norm}(cWx) = \frac{1}{c} \nabla_W \text{Norm}(Wx).
\end{equation}
As parameter norms grow during training, gradients diminish, reducing learning ability~\citep{lyle2024normalization,ceron2024in,palenicek2025scaling,mayor2025the,castanyer2025stable}. Moreover, uneven growth across layers causes inconsistent gradient scales, destabilizing optimization~\citep{lee2025simbav2}. 
This highlights the importance of controlling weight and gradient norms, in addition to the feature norm. Thus, we employ the following design choices to achieve stable norms.

We first opt LayerNorm as the forefront layer of both encoder and critic module, in order to control the norm of not only their intermediate features but also their inputs. While unusual for convolutional networks, this resembles the Dual PatchNorm design~\citep{kumar2023dual} which has been empirically shown to stabilize the gradients of the embedding layer~\footnote{In practice, we adopt the shift-and-norm strategy introduced in SimbaV2~\citep{lee2025simbav2} to preserve magnitude information. We use $\ell_2$-norm for action inputs however, as when $|\mathcal{A}|=1$ shift-and-LN always outputs $[-1,1]$.}. For preventing parameter growth, we surprisingly found a simple $\ell_2$ weight regularization to be sufficient, as shown in Figure~\ref{fig:metric_analysis}.

We further stabilize gradients by employing a distributional critic with KL divergence loss~\citep{bellemare2017distributional} and reward normalization~\citep{lee2025simbav2}. The KL divergence loss is more robust to noisy targets than mean squared error due to its smoother loss landscape~\citep{farebrother2024stop}, while reward normalization ensures consistent learning signals despite varying reward scales.

Specifically, reward normalization maintains unit variance in expected returns. Given reward $r_t$ at time $t$, we track the discounted return: 
\begin{equation}
    G_t \leftarrow \gamma G_{t-1} + r_t
\end{equation} 
with $G_t$ re-initialized to $0$ at the start of each episode. Let $\sigma^2_{t,G}$ denote the running variance of $G_t$. Each reward is then scaled as: 
\begin{equation}
    \bar{r}_t \leftarrow \frac{r_t}{\sqrt{\sigma_{t,G}^2 + \epsilon}},
\end{equation}

\subsection{Design Philosophy II: Maintaining Computational Efficiency}
\label{section:philosophy2}

Adding normalization layers and regularization increases training cost, so reducing computation is crucial. We find that most of DrQ-v2’s computational cost comes from early convolutional layers processing high-resolution inputs. We apply early downsampling via large-stride convolutions, a common practice in ResNet~\citep{he2016deep}, ConvNeXt~\citep{liu2022convnext}, and Vision Transformer~\citep{dosovitskiy2020image}. This results in an early reduction in spatial resolution and in turn, the computational cost of subsequent convolution layers.

We further cut computation by replacing convolutional layers into more cost-effective alternatives. For spatial convolutions, we instead utilize pointwise ($1 \times 1$ kernel) convolutions~\citep{hua2018pointwise}, which operate channel-wise without mixing spatial information, preserving spatial details at a lower cost. For downsampling, we adopt parameter-free $2 \times 2$ max-pooling~\citep{krizhevsky2012imagenet} in place of strided convolutions, which retain the strongest local activations while inducing significantly less overhead in both training and inference. Empirically, we found that these substitutions do not meaningfully alter the learning dynamics or the learning curves, indicating that the computational savings come at no cost to stability. 

\begin{figure}[t!]
\begin{center}
\includegraphics[width=0.9\textwidth]{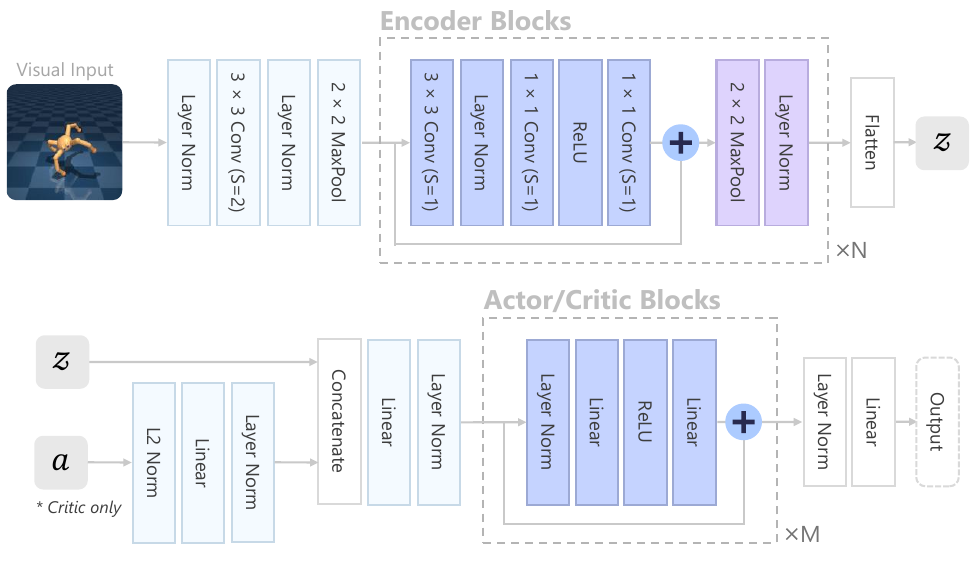}
\caption{\textbf{V-Simba architecture.} We aim to develop an architecture that constrains its feature norm, weight norm, and gradient norm for better stability and generalization. Precisely, we make extensive use of layer normalization and residual connections for stable features and gradients, and incorporate point-wise convolutions in tandem with spatial convolutions for computational efficiency.}
\label{fig:method}
\end{center}
\vspace{-0.3in} 
\end{figure}

\subsection{V-Simba Architecture}
\label{section:vsimba_arch}

Building on our design principles, we now detail the V-Simba architecture (Figure~\ref{fig:method}).

\textbf{Image Preprocessing.}
The input $o \in \mathbb{R}^{84 \times 84 \times 9}$ is a stack of the last three RGB frames. We first apply an alternating sequence of normalization and reduction steps. Specifically, the input passes through an initial LayerNorm, followed by a $3 \times 3$ convolution (stride 2), a second LayerNorm, and finally a $2 \times 2$ max-pooling layer (stride 2). This results in downsampled and normalized features $f_0 \in \mathbb{R}^{21 \times 21 \times 32}$.



\textbf{Encoder.}
The encoder consists of two sequential blocks transforming and downsampling features:
$f_0 \xrightarrow{\text{Block}_1} f_1 \xrightarrow{\text{Block}_2} f_2$,
where $f_1 \in \mathbb{R}^{10 \times 10 \times 32}, f_2 \in \mathbb{R}^{5 \times 5 \times 32}.$

Each encoder block processes input $f_i$ as follows:
\begin{enumerate}[leftmargin=2em,itemsep=0.5em, parsep=0pt, topsep=0.0em]
\item A $3 \times 3$ convolution to aggregate spatial features without changing resolution.
\item  Feature normalization (LayerNorm) to provide stable features to the subsequent layers.
\item Two pointwise ($1 \times 1$) convolutions with nonlinearities to refine and filter features, followed by a residual connection for stable gradient flow. Here, we employ an inverted bottleneck with 4$\times$ expansion, following ConvNext~\citep{liu2022convnext} and Simba~\citep{lee2024simba}, 
\item Downsampling with a $2 \times 2$ maxpool layer with stride 2 to reduce spatial resolution.
\item Applying LayerNorm to normalize the features before passing to the next block.
\end{enumerate}
After the second block, $f_2$ is flattened into the latent state vector $z \in \mathbb{R}^{800}$.

\textbf{Predictor.} The latent vector $z$ feeds into separate actor and critic heads, each followed by a linear layer and LayerNorm. For the critic, actions are separately embedded and concatenated with image embedding. The actor and critic embeddings have dimensions $z_\pi \in \mathbb{R}^{128}$, $z_Q \in \mathbb{R}^{512}$ following \cite{lee2024simba}.

Each embedding passes through residual nonlinear blocks: one block for the actor and two for the critic. Finally, the actor output passes through LayerNorm, a linear layer, and a $\tanh$ activation, while the critic output passes through LayerNorm and a linear layer modeling the Q-value distribution.

\begin{figure}[b]
    \centering
    \begin{minipage}{0.333\textwidth}
        \centering
        \begin{subfigure}{0.49\textwidth}
            \includegraphics[scale=0.21]{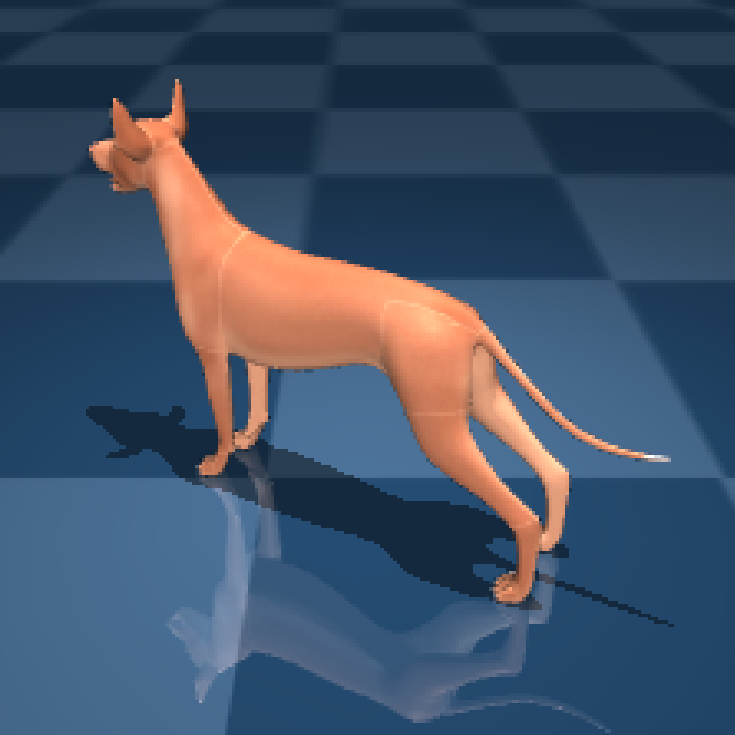}
        \end{subfigure}%
        \begin{subfigure}{0.49\textwidth}
            \includegraphics[scale=0.21]{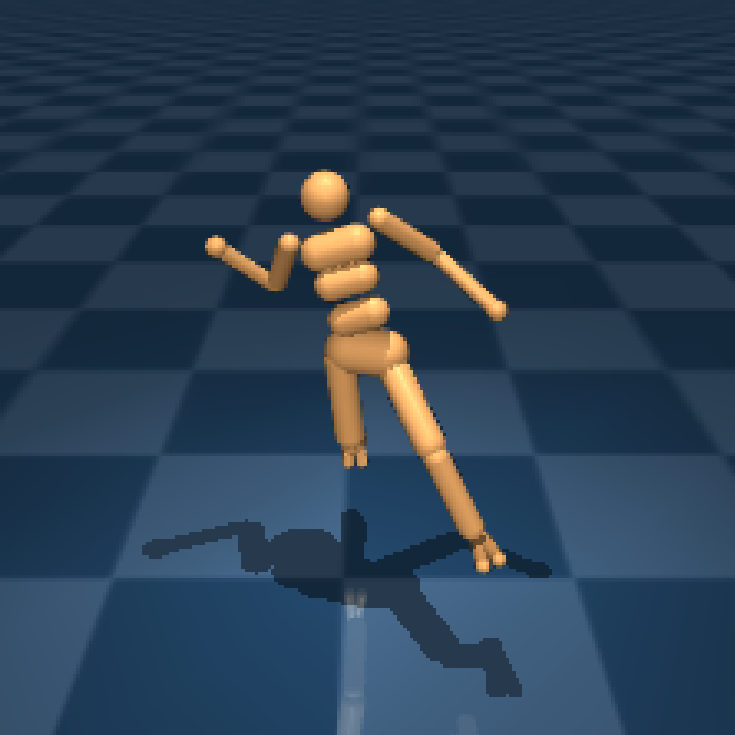}
        \end{subfigure}
        \subcaption{DeepMind Control Suite}
    \end{minipage}%
    \begin{minipage}{0.333\textwidth}
        \centering
        \begin{subfigure}{0.49\textwidth}
            \includegraphics[scale=0.21]{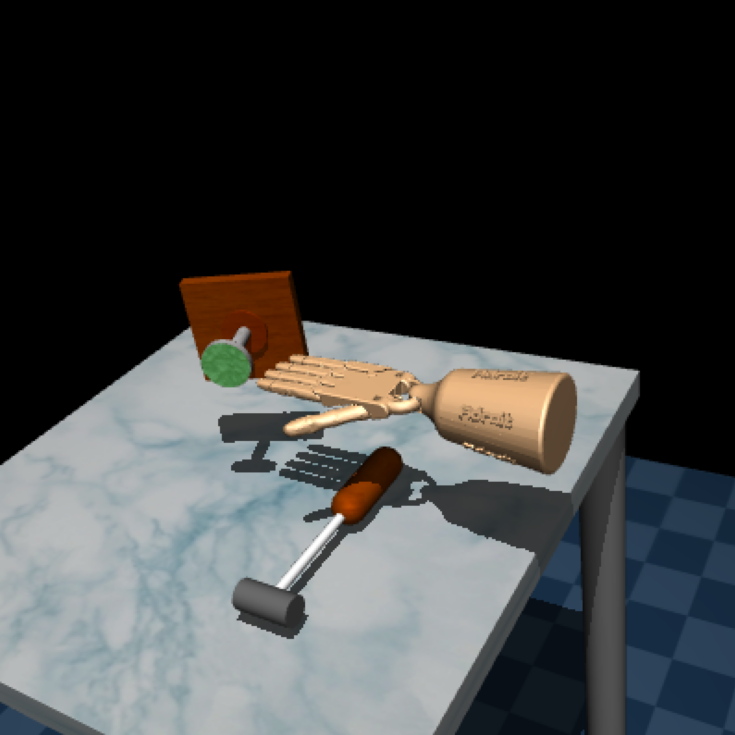}
        \end{subfigure}%
        \begin{subfigure}{0.49\textwidth}
            \includegraphics[scale=0.21]{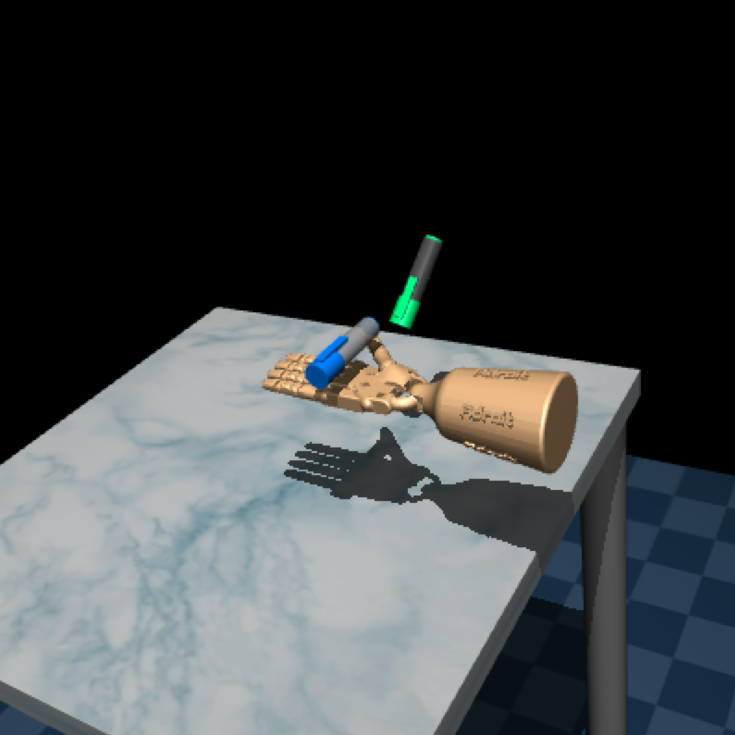}
        \end{subfigure}
        \subcaption{Adroit}
    \end{minipage}%
    \begin{minipage}{0.333\textwidth}
        \centering
        \begin{subfigure}{0.49\textwidth}
            \includegraphics[scale=0.21]{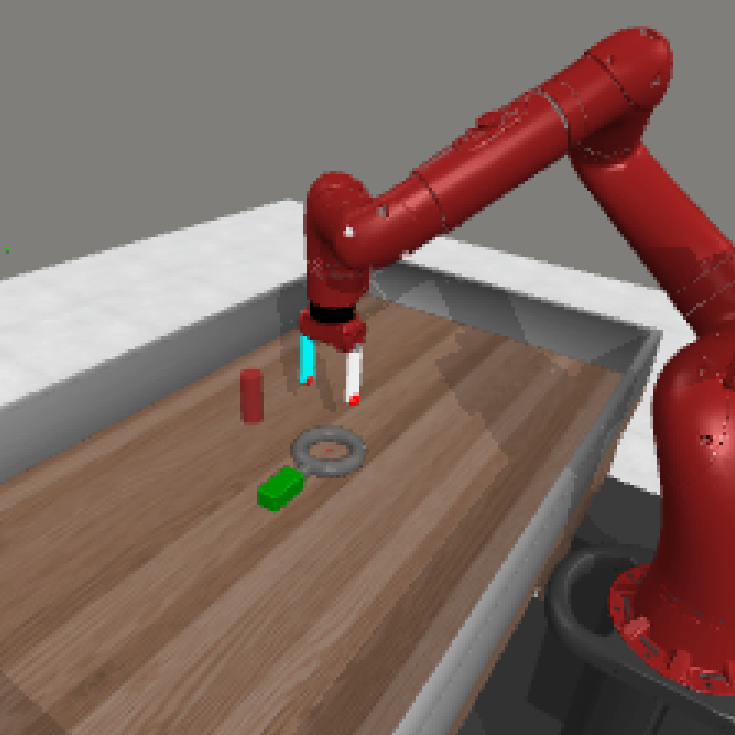}
        \end{subfigure}%
        \begin{subfigure}{0.49\textwidth}
            \includegraphics[scale=0.21]{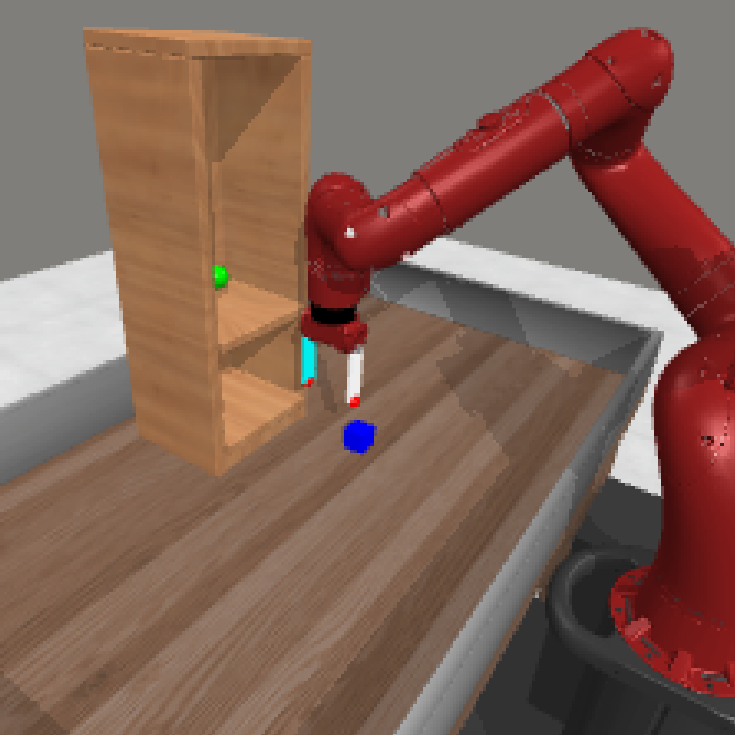}
        \end{subfigure}
        \subcaption{Meta-World}
    \end{minipage}
    \caption{\textbf{Environment Visualization.} We evaluate our V-Simba on 3 visual continuous control benchmarks: DeepMind Control Suite~\citep{tassa2018dmc},  Adroit~\citep{rajeswaran2017adroit}, and Meta-World~\citep{yu2020metaworld}.}
    \label{fig:environment_visualization}
\vspace{-1em}
\end{figure}

\section{Experiments}

We now provide an empirical evaluation of V-Simba:
\begin{enumerate}[leftmargin=2em,itemsep=0.5em, parsep=0pt, topsep=0.0em]
    \item \textbf{Performance Evaluation} (Sections~\ref{section:performance_evaluation}), comparing V-Simba against leading visual RL methods to demonstrate its effectiveness across diverse benchmarks.
    \item \textbf{Ablation Study} (Section~\ref{section:ablation_study}), conducting experiments demonstrating the contribution of each architectural component in V-Simba.
\end{enumerate}

\subsection{Experimental Setup}
\label{section:experimental_setup}

\textbf{Environment.} We consider a total of 29 continuous control tasks spanning 3 benchmarks: DeepMind Control (DMC) Suite~\citep{tassa2018dmc},  Adroit~\citep{rajeswaran2017adroit}, and Meta-World~\citep{yu2020metaworld}. Figure~\ref{fig:environment_visualization} shows the visualization of each task. These environments pose diverse challenges, including high-dimensional action spaces, sparse rewards, and complex dexterous manipulation, often under rich visual observations with shading and textures. Consequently, to solve the tasks, prior visual RL methods typically require either large volumes of frames or privileged information such as low-level robot states. Comprehensive descriptions of the benchmarks are provided in Appendix~\ref{appendix:environments}.

\textbf{Baselines.} In experiments, we compare V-Simba against a diverse set of state-of-the-art visual RL methods exemplifying three key algorithmic strategies: data and model regularization (DrQ-v2~\citep{yarats2021drqv2}, A-LIX~\citep{cetin2022alix}), advanced exploration (DrM~\citep{xu2023drm}), and model-based representation learning (TACO~\citep{zheng2023taco}, TD-MPC2~\citep{hansen2023tdmpcv2}, MR.Q~\citep{fujimoto2025mrq}). Notably, A-LIX, TACO, and DrM build upon DrQ-v2 (see Section~\ref{preliminary:drq}): A-LIX stabilizes training by adaptively regularizing the encoder’s gradients; TACO leverages a latent dynamics loss for richer representations; and DrM integrates dormant ratio~\citep{sokar2023redo}-guided mechanisms that balance exploration-exploitation dynamically. While these variants benefit from task-specific hyperparameter tuning, our method uses the \textit{same} hyperparameters across all tasks (see Appendix~\ref{appendix:hyperparameters} for the complete hyperparameter list). Whenever possible, we report original paper results; otherwise, we run the authors’ official implementations.

\subsection{Performance Evaluation}
\label{section:performance_evaluation}

\begin{figure}[t!]
\begin{center}
\includegraphics[width=1.0\textwidth]{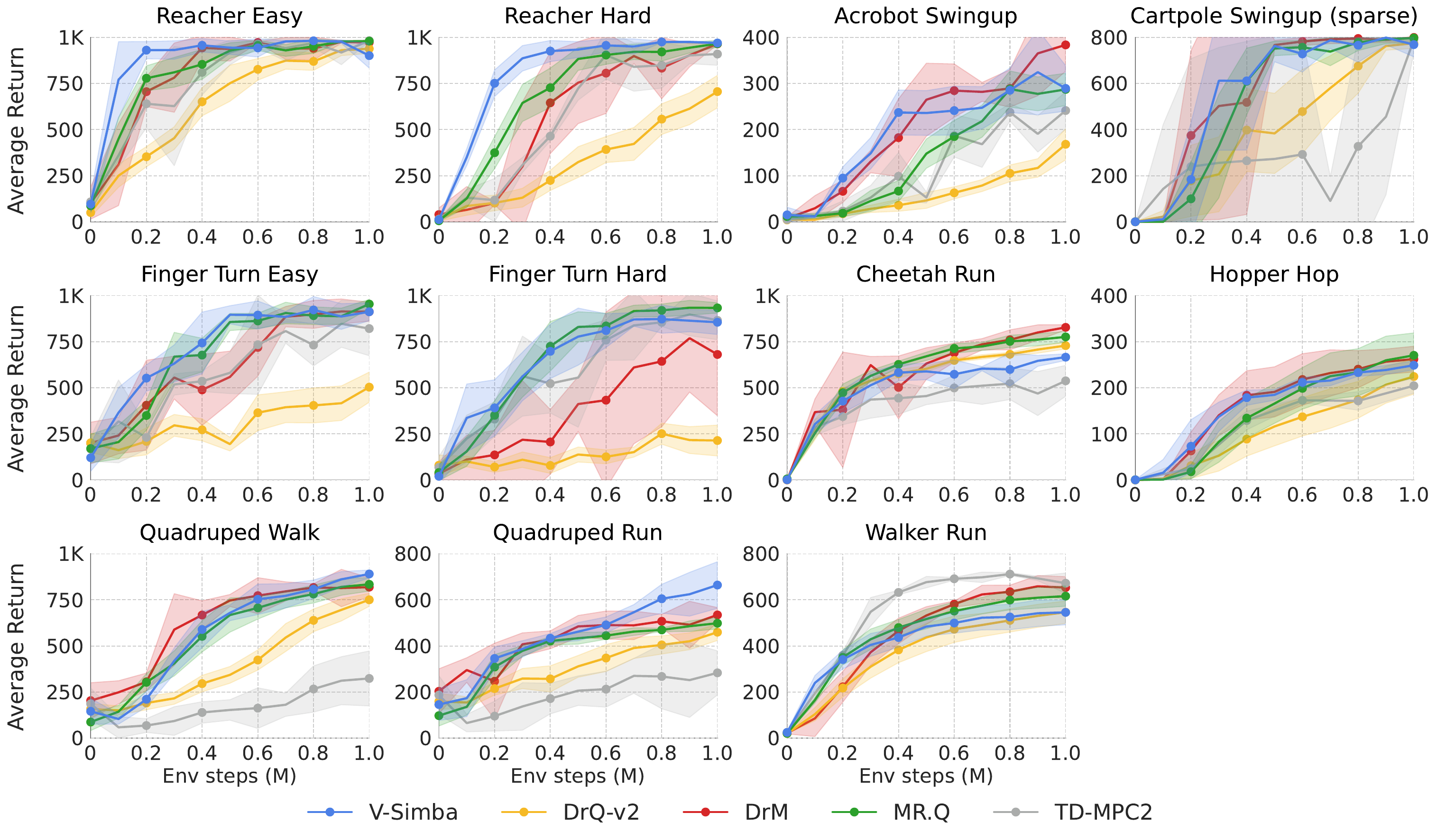}
\vspace{-0.7cm}
\caption{\textbf{DeepMind Control Suite - Medium.} Average episode returns on 11 medium-level tasks from DeepMind Control Suite~\citep{tassa2018dmc}. Each curve represents the mean performance across 5 random seeds per algorithm; shaded areas indicate 95\% bootstrap confidence intervals.}
\label{fig:performance_dmc_medium}
\end{center}
\vspace{-0.15in}
\end{figure}

\textbf{DMC Medium.} We begin by evaluating V-Simba on DMC Medium, consisting of 11 mid-difficulty tasks from DMC. As shown in Figure~\ref{fig:performance_dmc_medium}, our base algorithm, DrQ-v2, falls behind model-based methods such as TD-MPC2 and MR.Q. However, simply replacing DrQ-v2’s neural network with our proposed architecture, V-Simba, yields substantial performance gains. As a result, V-Simba surpasses TD-MPC2 and achieves results competitive with leading algorithm, MR.Q, highlighting the impact of architectural improvements.

\begin{figure}[t!]
\begin{center}
\includegraphics[width=1.0\textwidth]{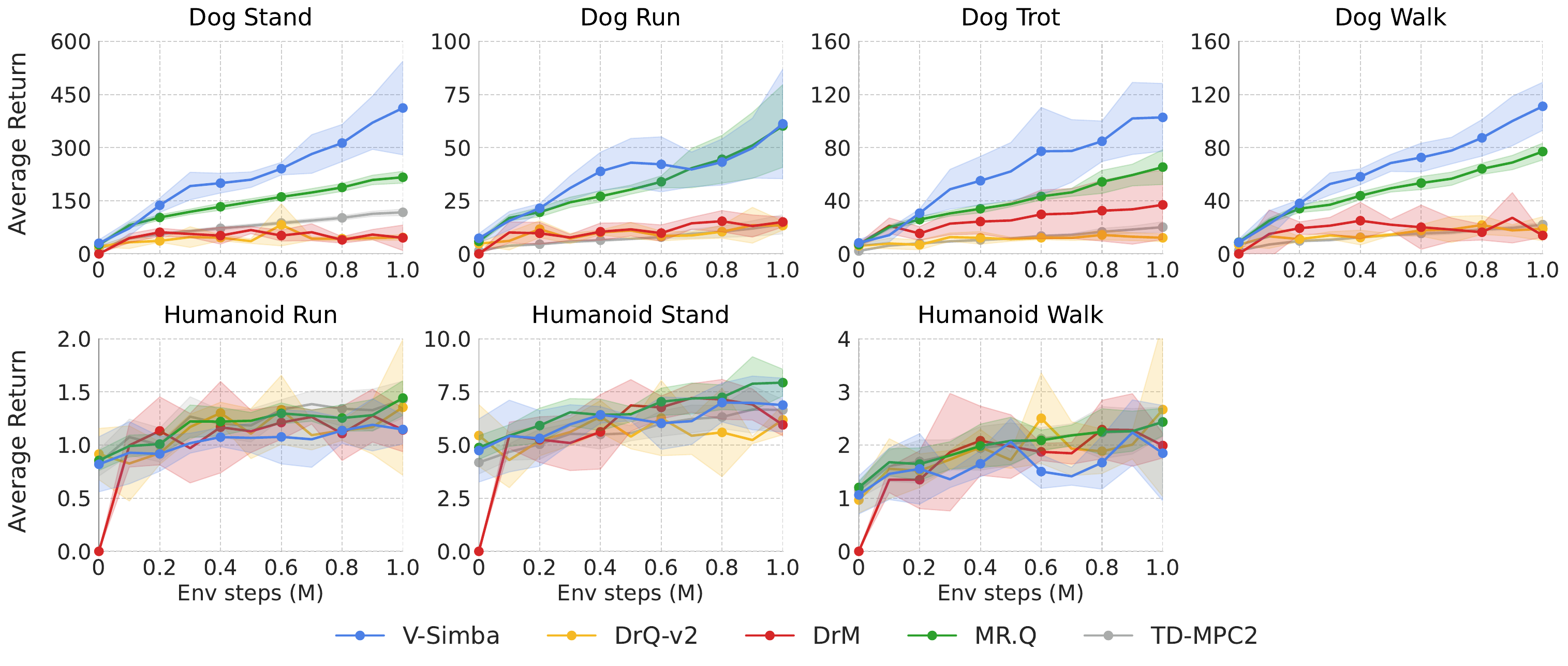}
\vspace{-0.7cm}
\caption{\textbf{DeepMind Control Suite - Hard.} Average episode returns on 7 hard-level tasks from DeepMind Control Suite~\citep{tassa2018dmc}. Each curve represents the mean performance across 5 random seeds per algorithm; shaded areas indicate 95\% bootstrap confidence intervals.}
\label{fig:performance_dmc_hard}
\end{center}
\vspace{-0.1in}
\end{figure}

\textbf{DMC Hard.} We further assess V-Simba on DMC Hard, a set of 7 high-difficulty tasks in DMC, characterized by complex kinematics and high-dimensional control. Figure~\ref{fig:performance_dmc_hard} shows that V-Simba performs competitively with MR.Q, though full task success remains elusive. We believe that this observation suggests that concurrent advances in both algorithm design and architectural representation are needed in visual RL to close the gap with state-based performance.

\begin{figure}[t!]
\begin{center}
\includegraphics[width=0.75\textwidth]{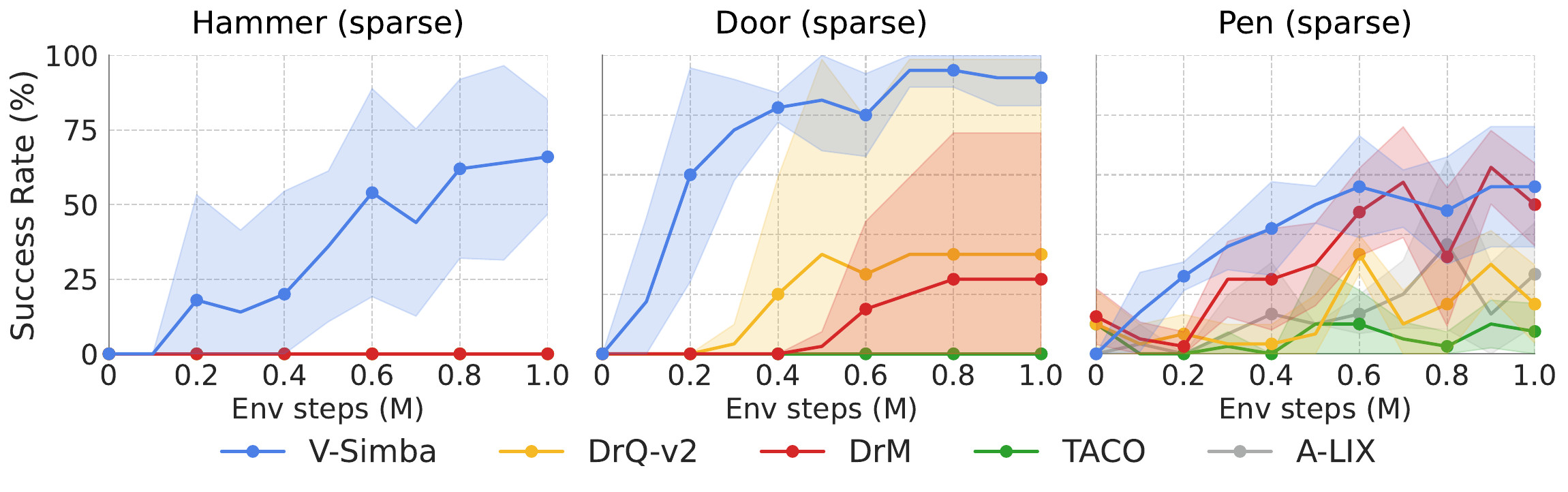}
\vspace{-0.3cm}
\caption{\textbf{Adroit - Sparse.} Average success rates on 3 sparse-reward tasks from Adroit~\citep{rajeswaran2017adroit}. Each curve represents the mean performance across 5 random seeds per algorithm; shaded areas indicate 95\% bootstrap confidence intervals.}
\label{fig:performance_adroit}
\end{center}
\vspace{-0.1in} 
\end{figure}
 
\begin{figure}[t!]
\begin{center}
\includegraphics[width=1.0\textwidth]{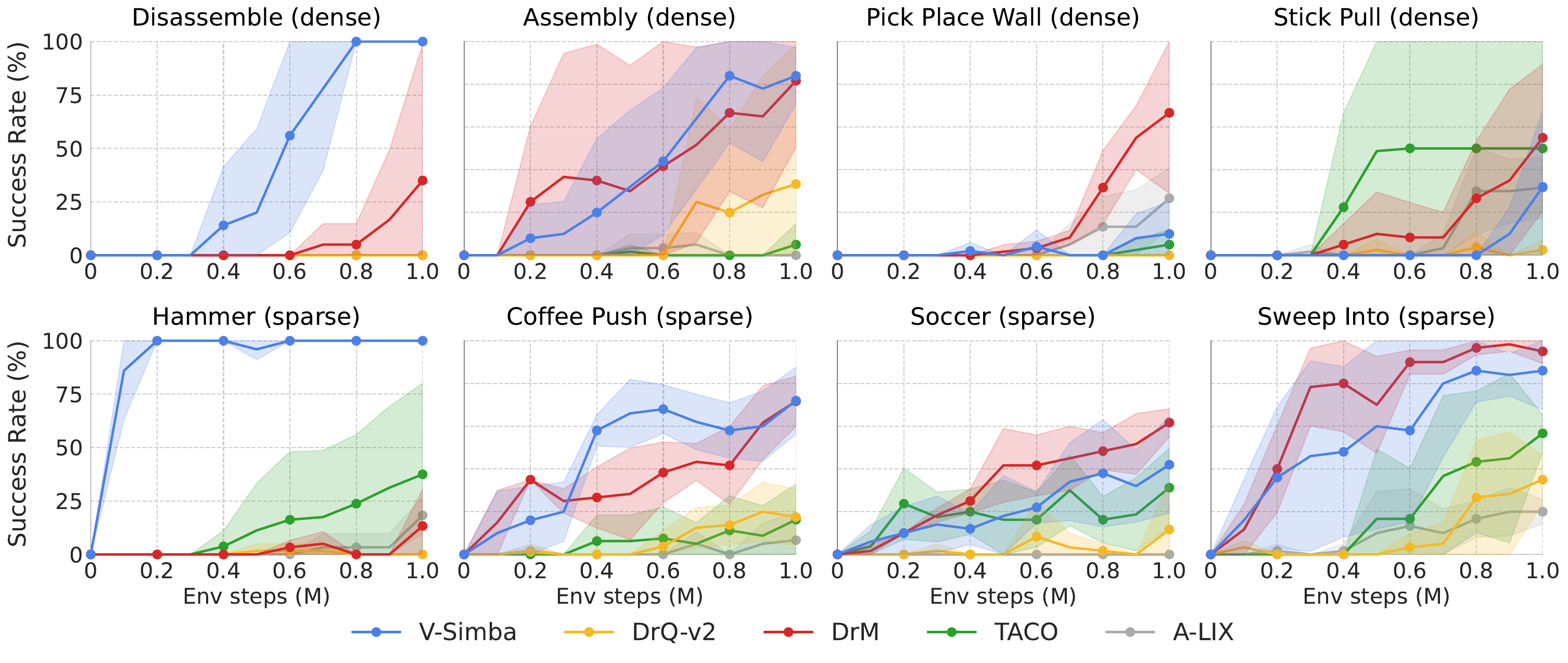}
\vspace{-0.7cm}
\caption{\textbf{Meta-World.} Average success rates on 8 tasks from the Meta-World~\citep{yu2020metaworld}. Each curve represents the mean performance across 5 random seeds per algorithm; shaded areas indicate 95\% bootstrap confidence intervals.}
\label{fig:performance_metaworld}
\end{center}
\vspace{-5mm} 
\end{figure}

\textbf{Adroit - Sparse.} Moving to more intricate scenarios, we evaluate V-Simba on Adroit under the challenging \textit{sparse-reward} setting. In this domain, the agent must control a dexterous hand-arm system to perform complex manipulation such as opening a door or using tools like a hammer. These tasks pose significant challenges for visual RL, often requiring over 5 million environment frames and access to privileged robot state inputs for successful learning. The comparison results are shown in Figure~\ref{fig:performance_adroit}. V-Simba reliably solves or nearly solves all tasks using only 1 million frames. In contrast, DrM--the previous state-of-the-art method--fails to learn meaningful behavior, despite leveraging privileged state vectors. Notably, V-Simba is the only method to solve \texttt{Hammer} with 1 million environment steps. These results underscore V-Simba's strong sample-efficiency and effectiveness in high-dimensional visual control settings. 


\textbf{Meta-World.} We also benchmark V-Simba on Meta-World, which demands precise object manipulation. We consider 4 medium-difficulty tasks: \texttt{Coffee Push}, \texttt{Soccer}, \texttt{Sweep Into}, and \texttt{Hammer}, and 4 high-difficulty tasks: \texttt{Assembly}, \texttt{Stick Pull}, \texttt{Pick Place Wall}, and \texttt{Disassemble}. For the medium tasks, we adopt a sparse-reward setting by replacing the ground-truth reward functions with binary success signals, following \cite{yu2020metaworld}, to increase task difficulty.  As shown in Figure~\ref{fig:performance_metaworld}, while DrQ-v2 struggles to learn in most tasks, V-Simba significantly improves over DrQ-v2 and matches or surpasses leading baselines, demonstrating superior sample efficiency. A notable performance improvement can be seen in the \texttt{Disassemble} and \texttt{Hammer} tasks, where V-Simba was able to consistently achieve almost perfect success rate, whereas prior works have failed in few trials.

\begin{table*}[ht]
\centering
\small
\caption{
\textbf{Ablation Study.} We exclude or modify each component in V-Simba and report their final performance on each benchmark, averaged over 3 random seeds. Each cell is highlighted base on their relative percentile difference to V-Simba, namely: positive \mgood{$(>0.01)$}, mildly negative \mbad{$[-0.05,-0.01)$}, damaging \mworse{$[-0.1,-0.05)$}, and catastrophic \mworst{$[-1.0,-0.1)$}.
}
\label{table:ablation}
\resizebox{\linewidth}{!}{%
\begin{tabular}{%
    @{}>{\raggedright\arraybackslash}m{4.2cm}  
    *{5}{>{\arraybackslash}m{2.2cm}@{\hspace{0.55cm}}}  
    *{1}{>{\arraybackslash}m{2.1cm}}}
\hline
\noalign{\vspace{1.5pt}}
\multirow{2}{*}{\textbf{Ablation}}
  & \textbf{DMC (18)}
  & \textbf{Adroit (3)}
  & \textbf{MetaWorld (8)}
  & \textbf{All (29)} \\

  & \textcolor{darkgray}{Return (1k)}
  & \textcolor{darkgray}{Success Rate}
  & \textcolor{darkgray}{Success Rate}
  & \textcolor{darkgray}{-} \\

\hline
\noalign{\vspace{1.5pt}}

\textbf{Normalization Layers}\\

(a) No Normalization Layers
 & \cellcolor{ab_worse} \po0.442 \scriptsize{\textcolor{gray}{$\pm$ 0.095\po}}
 & \cellcolor{ab_bad} \po0.694 \scriptsize{\textcolor{gray}{$\pm$ 0.122\po}}
 & \cellcolor{ab_worst} \po0.497 \scriptsize{\textcolor{gray}{$\pm$ 0.167\po}}
 & \cellcolor{ab_worst} \po0.492 \scriptsize{\textcolor{gray}{$\pm$ 0.081\po}}
 \\
(b) No Input Normalization
 & \cellcolor{ab_worst} \po0.401 \scriptsize{\textcolor{gray}{$\pm$ 0.098\po}}
 & \cellcolor{ab_worst} \po0.612 \scriptsize{\textcolor{gray}{$\pm$ 0.133\po}}
 & \cellcolor{ab_bad}\po0.642 \scriptsize{\textcolor{gray}{$\pm$ 0.158\po}}
 & \cellcolor{ab_worse} \po0.502 \scriptsize{\textcolor{gray}{$\pm$ 0.083\po}}
 \\
(c) LN $\rightarrow$ LN w/o $\gamma, \beta$
 & \cellcolor{ab_bad} \po0.453 \scriptsize{\textcolor{gray}{$\pm$ 0.073\po}}
 & \cellcolor{ab_worse} \po0.668 \scriptsize{\textcolor{gray}{$\pm$ 0.133\po}}
 & \cellcolor{ab_worst} \po0.586 \scriptsize{\textcolor{gray}{$\pm$ 0.125\po}}
 & \cellcolor{ab_bad} \po0.522 \scriptsize{\textcolor{gray}{$\pm$ 0.063\po}}
 \\

\noalign{\vspace{1.5pt}}
\hline
\noalign{\vspace{1.5pt}}
\textbf{Residual and Weight Decay} \\

(d) No Residual Connection
 & \cellcolor{ab_bad} \po0.453 \scriptsize{\textcolor{gray}{$\pm$ 0.098\po}}
 & \cellcolor{ab_worst} \po0.623 \scriptsize{\textcolor{gray}{$\pm$ 0.178\po}}
 & \cellcolor{ab_worse} \po0.618 \scriptsize{\textcolor{gray}{$\pm$ 0.154\po}}
 & \cellcolor{ab_bad} \po0.526 \scriptsize{\textcolor{gray}{$\pm$ 0.081\po}}
 \\
(e) No Weight Decay
 & \cellcolor{ab_bad} \po0.452 \scriptsize{\textcolor{gray}{$\pm$ 0.099\po}}
 & \cellcolor{ab_worst} \po0.591 \scriptsize{\textcolor{gray}{$\pm$ 0.144\po}}
 & \cellcolor{ab_worst} \po0.513 \scriptsize{\textcolor{gray}{$\pm$ 0.171\po}}
 & \cellcolor{ab_worst} \po0.492 \scriptsize{\textcolor{gray}{$\pm$ 0.080\po}}
 \\

\noalign{\vspace{1.5pt}}
\hline
\noalign{\vspace{1.5pt}}
\textbf{Value Learning} \\
 
(f) No Categorical Critic
 & \cellcolor{ab_worse} \po0.441 \scriptsize{\textcolor{gray}{$\pm$ 0.096\po}}
 & \cellcolor{ab_worst} \po0.591 \scriptsize{\textcolor{gray}{$\pm$ 0.144\po}}
 & \cellcolor{ab_worst} \po0.576 \scriptsize{\textcolor{gray}{$\pm$ 0.167\po}}
 & \cellcolor{ab_worse} \po0.504 \scriptsize{\textcolor{gray}{$\pm$ 0.079\po}}
 \\
(g) No Reward Scaling
 & \cellcolor{ab_worse} \po0.439 \scriptsize{\textcolor{gray}{$\pm$ 0.094\po}}
 & \cellcolor{ab_worse} \po0.683 \scriptsize{\textcolor{gray}{$\pm$ 0.144\po}}
 & \cellcolor{ab_worst} \po0.599 \scriptsize{\textcolor{gray}{$\pm$ 0.158\po}}
 & \cellcolor{ab_worse} \po0.519 \scriptsize{\textcolor{gray}{$\pm$ 0.080\po}}
 \\

\noalign{\vspace{1.5pt}}
\hline
\noalign{\vspace{1.5pt}}

V-Simba
  & \po 0.467 \scriptsize{\textcolor{gray}{$\pm$ 0.075\po}} 
  & \po 0.725 \scriptsize{\textcolor{gray}{$\pm$ 0.167\po}} 
  & \po 0.668 \scriptsize{\textcolor{gray}{$\pm$ 0.115\po}} 
  & \po 0.549 \scriptsize{\textcolor{gray}{$\pm$ 0.060\po}}
  \\

\noalign{\vspace{1pt}}
\hline

\end{tabular}
} 
\vspace{-1.2mm}
\end{table*}

\subsection{Ablation Study}
\label{section:ablation_study}

To assess the impact of each component on V‑Simba’s overall performance, we evaluate variants that remove or modify one component at a time. The results are reported in Table~\ref{table:ablation}.

We first investigate the effect of normalization layers (Table~\ref{table:ablation}.(a)-(c)). \textbf{No Normalization Layers} removes LayerNorm entirely from the network, whereas \textbf{No Input Normalization} only removes two LayerNorms: for image and action inputs in encoder and critic respectively. \textbf{LN w/o $\gamma, \beta$} removes the bias and scale parameters of LayerNorm. In summary, by removing certain normalization layers or components, the network loses control over the feature norms, leading to degradation.

Next, we quantify the importance of residual connections and weight decay (Table~\ref{table:ablation}.(d)-(e)). Both residual connections and weight decay, along with their well-known benefits, are also known to bias the network towards simple solutions for improved robustness~\citep{teney2024neuralredshift, lee2024simba}. Removing such components led to a visible drop in performance, similar to removing normalization layers.

Finally, the categorical critic and reward scaling are critical components, as they reformulate the regression problem into a categorical prediction, giving a much more stable gradient and learning dynamics. Reverting back to regression loss led to diminished performance (Table~\ref{table:ablation}.(f)). Even with categorical loss, leaving no bounds to the reward scales led to similar consequences (Table~\ref{table:ablation}.(g)), highlighting the importance of assuring the Q-values to stay in a certain range.

\section{Lessons and Opportunities}

In this work, we introduce V-Simba, a simple yet effective neural network architecture for visual continuous control, inspired by the Simba architecture from state-based RL~\citep{lee2024simba}. By combining feature normalization, weight regularization, and a distributional critic, V-Simba achieves superior performance over prior visual RL methods across multiple benchmarks with minimal algorithmic changes. Additionally, V-Simba reduces computational cost by integrating early downsampling through large-stride convolutions and pointwise convolution layers, enabling faster training than DrQ-v2~\citep{yarats2021drqv2}. We believe our work does not oppose the current trend of adopting model-based learning or exploration strategies; rather, it offers a complementary approach that can be integrated with subsequent studies.

Regardless, several opportunities remain. Because V-Simba constrains the growth of features, weights, and gradients, it offers a natural substrate for scaling to larger widths and depths, a promising avenue we leave to future work. Seed variance can also be large on a few tasks (e.g., \texttt{Adroit-Hammer}), though comparable spread arises across visual-RL methods rather than being specific to V-Simba, and reducing it remains important. Finally, our evaluation protocol focuses on evaluating wide range of environments with a single hyperparameter set with 5 seeds; more seeds or longer horizons would further sharpen these estimates.

In recent years, reinforcement learning for robotic control has gained increased attention. However, limited sample efficiency remains a significant barrier to real-world adoption. While simulators provide valuable virtual environments~\citep{makoviychuk2021isaac, zakka2025mujoco}, rendering high-resolution images with complex object interactions is still computationally expensive and difficult to parallelize. This underscores the importance of improving sample efficiency. V-Simba offers a lightweight architectural solution using well-established components that are easy to integrate into existing algorithms. Its simplicity allows practitioners to adopt and extend it with minimal overhead. We hope V-Simba serves as an architectural foundation to accelerate progress in the robotics community.

\subsubsection*{Acknowledgments}
\label{sec:ack}
This work was supported by Institute for Information \& Communications Technology Planning \& Evaluation(IITP) grant funded by the Korea government(MSIT) (RS-2019-II190075, Artificial Intelligence Graduate School Program(KAIST)). This work was supported by the National Research Foundation of Korea(NRF) grant funded by the Korea government(MSIT) (No. RS-2025-00555621). This research was supported by the “Advanced GPU Utilization Support Program” funded by the Government of the Republic of Korea (Ministry of Science and ICT).

Pablo Samuel Castro acknowledges funding from NSERC Discovery Grant. Aaron Courville acknowledge funding from National Sciences and Engineering Council of Canada (NSERC) and the Canadian Institute for Advanced Research (CIFAR). Johan Obando-Ceron acknowledge funding support from Google and CIFAR AI. We would also like to thank the Python community \citep{van1995python, 4160250} for developing tools that enabled this work, including NumPy \citep{harris2020array}, Matplotlib \citep{hunter2007matplotlib}, Jupyter \citep{2016ppap}, and Pandas \citep{McKinney2013Python}.







\bibliography{main}
\bibliographystyle{rlj}

\beginSupplementaryMaterials

\section{Extended Metric Analysis}
\label{appendix:metric_analysis}

This section provides the details and extended results of the metric analysis presented in Figure~\ref{fig:metric_analysis}.

\subsection{Setup}

Our goal is to analyze and compare neural network architectures in visual RL in terms of learning dynamics and stability. We use two baseline algorithms---DDPG~\cite{lillicrap2015ddpg} and SAC~\cite{haarnoja2018sac}---both with data augmentation~\cite{yarats2021drqv2}. We then evaluate four neural architectures proposed by DrQ-v2~\cite{yarats2021drqv2}, IMPALA~\cite{espeholt2018impala}, MR.Q~\cite{fujimoto2025mrq} and our proposed V-Simba on each algorithm. For IMPALA, we only employ the encoder with residual blocks, combined with DrQ-v2 predictors. For MR.Q, we exclude the dynamics learning components and learn the encoder and critic end-to-end (dubbed `no MR' in their ablation experiments).

We follow the original paper for any architecture-specific hyperparameters such as the number of layers, hidden dimension, and the use of clipped double Q-learning (CDQ)~\cite{fujimoto2018td3}. Otherwise, we use the same set of hyperparameters for all experiments. We measure the metrics (Section~\ref{appendix:metrics}) every 10,000 update steps (20,000 environment steps), using a mini-batch of size 256. 

\subsection{Metrics}
\label{appendix:metrics}

We employ the following metrics for analysis:

\textbf{Sharpness of the loss landscape.} Sharpness is often considered indicative of a neural network's ability to generalize. In reinforcement learning, the underlying data distribution is inherently non-stationary, making consistent generalization crucial. We quantify sharpness by the largest eigenvalue of the Hessian matrix ($\lambda_{\max}(\nabla^2\mathcal{L})$)~\cite{lee2024plastic,foret2020sharpness}, which can be approximated using the Lanczos algorithm~\citep{golub1996lanczos}.

\textbf{Dormant ratio.} A neuron is said to be inactive or \textit{dormant} when its absolute activation value tends to be small compared to the layer's average. Formally, the $i$-th neuron of layer $\ell$ is $\tau$-dormant if $s^\ell_i=\frac{\mathbb{E}_{x\in D}|h^\ell_i(x)|}{\frac{1}{H^\ell}\sum_{k\in h}\mathbb{E}_{x\in D}|h^\ell_i(x)|}\leq\tau$, where $h^\ell_i$ are the activation values of layer $\ell$~\cite{sokar2023redo}. A high proportion of dormant neurons implies that the network's decisions rely heavily on only a few neurons, indicating capacity loss. We use $\tau=0.1$ in our analysis.

\textbf{Feature diversity.} While maximizing feature diversity itself might not be crucial for RL, preventing feature collapse is critical, as it reduces the network's capacity and hampers learning capability. Inspired by ConvNext-v2~\cite{woo2023convnextv2}, we measure the average cosine distance between the samples within a batch: $\frac{1}{B^2}\sum^B_i\sum^B_j \frac{1-cos(X_i,X_j)}{2}$, where $B$ is the batch size, and $X\in\mathbb{R}^{B\times D}$ is the feature matrix.

\textbf{Norms.} Prior works in state-based RL have shown that controlling the growth of features, weights and gradient norms can stabilize the learning process and thus performance~\cite{lee2024simba, lee2025simbav2, palenicek2025scaling}. We investigate whether the same argument could be made for visual RL as well. Following \cite{lee2025simbav2}, we define the \textit{effective} norm of a set of vectors and matrices using dimension-based weights $w_i (z) = \frac{\mathrm{dim} (z_i)}{\sum_{j=1}^N \mathrm{dim} (z_j)}$, which captures dimensional contributions across vectors and matrices. For example, for a neural network's parameter set $\theta = \{ \theta_i \}_{i=1}^N$, the effective parameter norm is defined as $\Vert \theta \Vert_{\mathrm{eff}}^2 \triangleq \sum_{i=1}^N w_i (\theta) \Vert \theta_i\Vert_2$ where $\Vert \cdot \Vert_2^2$ denotes the standard $\ell_2$-norm (or Frobenius norm $\Vert \cdot \Vert_F$ for matrices). 

\subsection{Results}

\begin{figure}[t!]
\begin{center}
\includegraphics[width=1.0\textwidth]{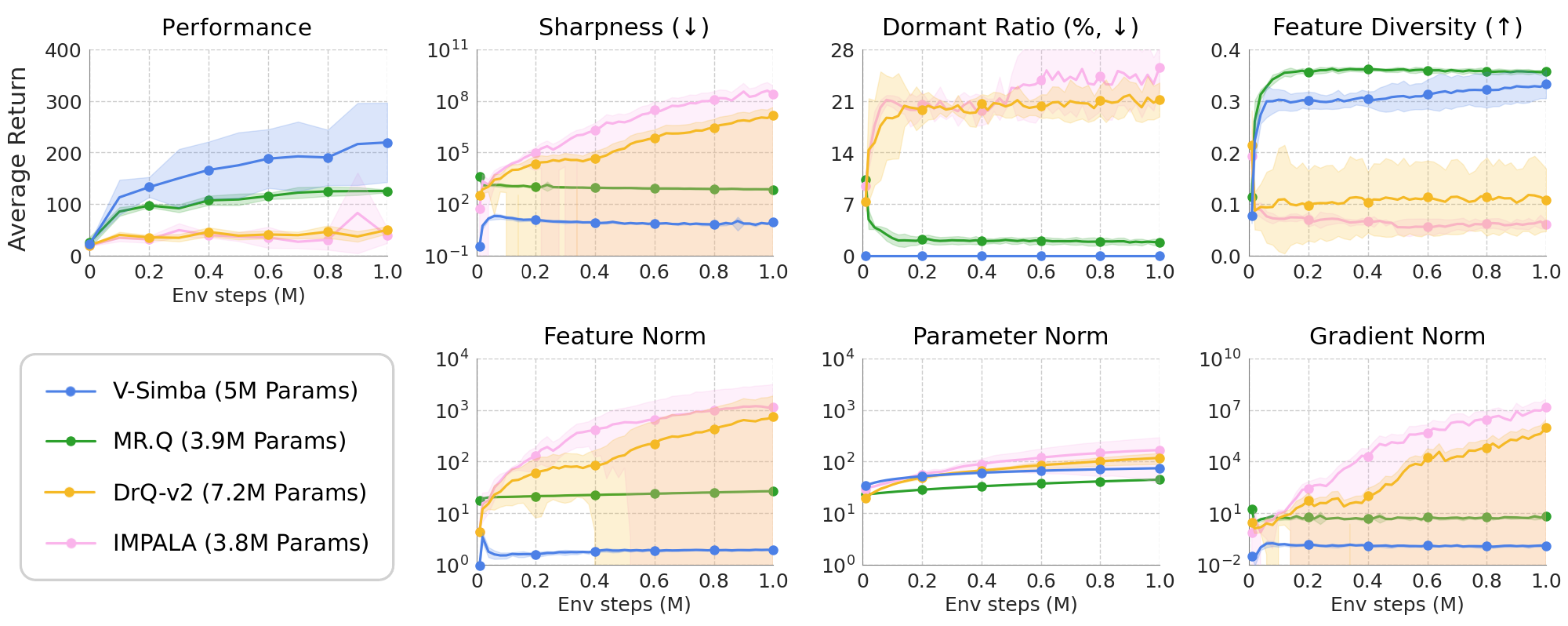}
\vspace{-2mm}
\caption{\textbf{Comparison of Neural Architectures under DDPG.} We evaluate and compare the neural network architectures proposed by DrQ-v2, IMPALA, MR.Q and our V-Simba, using DDPG with data augmentation in the \texttt{Dog Stand} environment. V-Simba maintains greater stability throughout training and outperforms other baselines.}
\label{fig:metric_analysis_extended_ddpg}
\end{center}
\vspace{-0.2in} 
\end{figure}

\begin{figure}[t!]
\begin{center}
\includegraphics[width=1.0\textwidth]{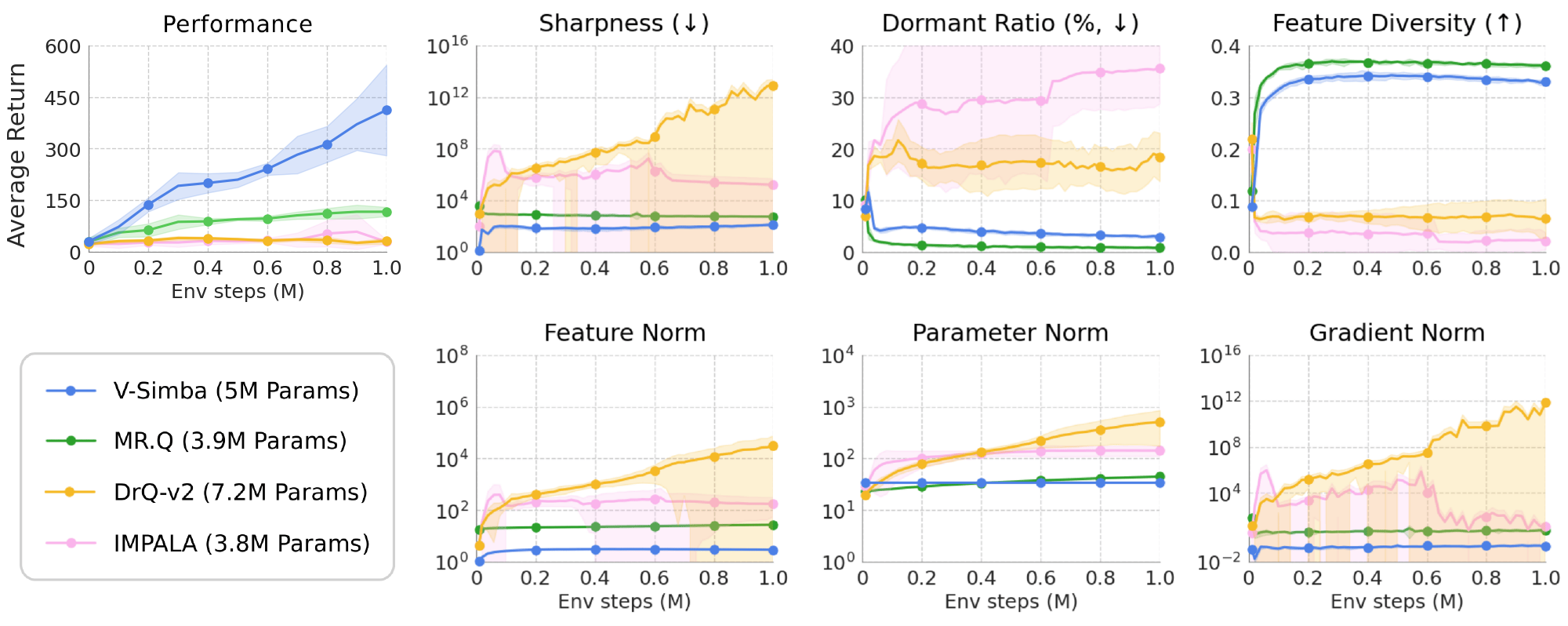}
\vspace{-2mm}
\caption{\textbf{Comparison of Neural Architectures under SAC.} We evaluate and compare the neural network architectures proposed by DrQ-v2, IMPALA, MR.Q and our V-Simba, using SAC with data augmentation in the \texttt{Dog Stand} environment. V-Simba maintains greater stability throughout training and outperforms other baselines.}
\label{fig:metric_analysis_extended_sac}
\end{center}
\vspace{-0.2in} 
\end{figure}

We visualize the results for DDPG in Figure~\ref{fig:metric_analysis_extended_ddpg}, and SAC in Figure~\ref{fig:metric_analysis_extended_sac}. DrQ-v2 and IMPALA architectures exhibit significant instability across both algorithms, showing high sharpness and dormant ratio, low feature diversity, and exploding feature, parameter and gradient norms. Collectively, these issues hinder the learning process and their capacity to learn meaningful behaviors.

Meanwhile, MR.Q maintains better stability by incorporating numerous normalization layers into its design. Notably, MR.Q achieves lower dormant ratios and higher feature diversity compared to V-Simba, highlighting the importance of normalization layers in stabilizing learning dynamics. Despite the strengths, MR.Q still experiences relatively high sharpness and norm magnitudes, although their growth is better controlled.

Finally, our proposed V-Simba further enhances stability by rigorously controlling the norm scales. This leads to smoother loss landscape, low dormant ratio, high feature diversity, and overall superior performance compared to all other evaluated methods.

We stress that these metrics are diagnostic rather than predictive: no single metric cleanly orders methods by return. MR.Q, for instance, matches or exceeds V-Simba on dormant ratio and feature diversity, yet attains a lower average return, meaning low dormancy or high diversity alone is not sufficient for strong performance. Among the metrics we track, sharpness aligns most consistently with the observed performance ordering, and MR.Q's comparatively high sharpness may partially account for its gap to V-Simba, in line with the established connection between sharpness and generalization~\citep{foret2020sharpness, lee2024plastic}. Accordingly, we treat these metrics as complementary lenses on optimization stability that motivate our architectural choices, rather than as standalone predictors of performance.

\section{Hyperparameters}
\label{appendix:hyperparameters}

Table~\ref{table:hyperparameters} lists the hyperparameters used across all experiments. Unless otherwise specified, we follow the default hyperparameters from prior work for consistency and computational efficiency, despite the well-known sensitivity of deep RL agents to these choices~\citep{ceron2024on}.

\vspace{-2.5mm}%
\begin{table}[ht]
\centering
\caption{\textbf{Hyperparameters Table.} We use the consistent hyperparameters across all benchmarks, which are listed below. The discount factor $\gamma$ is automatically determined by heuristics from \citep{hansen2023tdmpcv2}.}
\small
\vspace{4mm}
\label{table:hyperparameters}
\resizebox{0.90\textwidth}{!}{

\begin{tabular}{llll}
\hline
\noalign{\vspace{1.5pt}}
& \textbf{Hyperparameter} & \textbf{Notation} & \textbf{Value} \\
\noalign{\vspace{1.5pt}}
\hline
\noalign{\vspace{1.5pt}}
\multirow{6}{*}{\textbf{Common}}
& Discount factor & $\gamma$ & Heuristic~\citep{hansen2023tdmpcv2} \\
& Replay buffer capacity     & - & $1$M \\
& Buffer sampling            & - & Uniform \\
& Batch size                 & - & $256$ \\
& Observation Shape & $\vert \mathcal{O} \vert$ & $3 \times 84 \times 84$ \\
& Update-to-data (UTD) ratio & - & $1$ \\
& TD steps                   & $n$ & $3$ \\
\noalign{\vspace{1.5pt}}
\hline
\noalign{\vspace{1.5pt}}
\multirow{2}{*}{\textbf{Encoder}} 
& Number of blocks & $L$ & $2$ \\
& Hidden dimension (channels) & $d_h$ & $32$ \\
\noalign{\vspace{1.5pt}}
\hline
\noalign{\vspace{1.5pt}}
\multirow{4}{*}{\textbf{Predictor - Actor}} 
& Number of blocks & $L$ & 1 \\
& Hidden dimension & $d_h$ & 128 \\
& Initial temperature & $\alpha_0$ & $1\mathrm{e}{-2}$
\\
& Target entropy & $\mathbb{H}^*$ & $\vert\mathcal{A}\vert/2$ \\
\noalign{\vspace{1.5pt}}
\hline
\noalign{\vspace{1.5pt}}
\multirow{6}{*}{\textbf{Predictor - Critic}} 
& Number of blocks & $L$ & 2 \\
& Hidden dimension & $d_h$ & 512 \\
& Number of atoms  & $n_\text{atoms}$ & 101 \\
& Target critic momentum & $\tau$ & $5\mathrm{e}{-3}$
 \\ 
& Clipped double Q               & - & No \\
& Action embedding dimension     & $d_a$ & 128 \\
\noalign{\vspace{1.5pt}}
\hline
\noalign{\vspace{1.5pt}}
\multirow{3}{*}{\textbf{Output}} 
& Number of return bins & $n_\text{atoms}$ & $101$ \\
& Support of return & $[G_{\min}, G_{\max}]$ & $[-5, 5]$ \\
& Reward scaler epsilon & $\epsilon$ & $1\mathrm{e}{-8}$ \\
\noalign{\vspace{1.5pt}}
\hline
\noalign{\vspace{1.5pt}}
\multirow{4}{*}{\textbf{Optimizer}} 
& Optimizer          & - & Adam \\
& Optimizer momentum & $(\beta_1, \beta_2)$ & (0.9, 0.999) \\
& Weight Decay       & - & 1$\mathrm{e}{-2}$ \\
& Learning rate & $\eta$ & $1\mathrm{e}{-4}$ \\ 
\noalign{\vspace{1.5pt}}
\hline
\end{tabular}
    }
\end{table} %

\section{Compute Resources}

We mainly use RTX3090 GPUs in our experiments, which takes approximately $4.8$ hours to finish a single seed experiment with V-Simba. We have optimized the replay buffer to be more memory-friendly, requiring around 11GB of RAM memory for each experiment.


\section{Baselines}
\label{appendix:baselines_online}

\textbf{DrQ-v2}~\citep{yarats2021drqv2}. Data-regularized Q-learning (DrQ-v2) incorporates data augmentation via random shift transformations into DDPG to avoid overfitting of visual encoder to specific visual patterns. We provide a detailed explanation of the algorithm in Section~\ref{preliminary:drq}. Results for DMC Medium and Meta-World were obtained from \citep{yarats2021drqv2} and \citep{xu2023drm}, respectively. We run the official repository (\url{https://github.com/facebookresearch/drqv2}) over 3 random seeds for Adroit, and 5 random seeds for DMC Hard results.

\textbf{A-LIX}~\citep{cetin2022alix}. Adaptive Local Signal Mixing (A-LIX) modifies the convolutional layer of DrQ-v2 by performing bilinear interpolation with weights derived from random shifts, regularizing gradients and reducing overfitting. Meta-World results are from \citep{xu2023drm}, which are averaged over 4 random seeds. We obtained Adroit results by running the official repository (\url{https://github.com/Aladoro/Stabilizing-Off-Policy-RL}) over 3 seeds.

\textbf{TACO}~\citep{zheng2023taco}. Temporal Action-driven Contrastive Learning (TACO) jointly learns state and action representations introducing contrastive learning to DrQ-v2, which promotes to generalize its knowledge more effectively across diverse state-action pairs, enhancing the sample efficiency of RL algorithms. Meta-World results are from \citep{xu2023drm}, which are averaged over 4 random seeds. We obtained Adroit results by running the official repository (\url{https://github.com/FrankZheng2022/TACO}) over 3 seeds. 

\textbf{DrM}~\citep{xu2023drm}. Dormant Ratio Minimization (DrM) extends DrQ-v2 with three mechanisms that reduce the agent’s dormant ratio and leverage it to balance exploration and exploitation. Meta-World results are from \citep{xu2023drm}, which are averaged over 4 random seeds. We obtained Adroit results by running the official repository (\url{https://github.com/XuGW-Kevin/DrM}) over 3 seeds.

\textbf{DreamerV3}~\citep{hafner2023dreamerv3}. DreamerV3 builds a latent world model by encoding the observation into a compact latent space for long-horizon behavior and value learning. DMC results are from \citep{fujimoto2025mrq}, reproduced with the official codebase (\url{https://github.com/danijar/dreamerv3}) over 10 seeds.

\textbf{TD-MPC2}~\citep{hansen2023tdmpcv2}. TD-MPC2 learns a decoder-free world model via multi-task dynamics prediction and performs latent-space planning. DMC results are from \citep{fujimoto2025mrq}, reproduced with the official codebase (\url{https://github.com/nicklashansen/tdmpc2}) over 10 seeds.

\textbf{MR.Q}~\citep{fujimoto2025mrq}. Model-based Representations for Q-learning (MR.Q) is a model-free algorithm that leverages model-based auxiliary tasks to learn rich actor-critic representations. DMC results are from \citep{fujimoto2025mrq}, averaged over 10 seeds.


\clearpage
\section{Environment Details}
\label{appendix:environments}

This section describes the benchmark environments used in our evaluation.  A complete list of tasks, including state and action dimensions, is provided at the end of the section. Although state vectors are not used during training, we report them to reflect task difficulty.  Table~\ref{tab:appendix_environment_details} details episode length, frame stack, action repeat, total environment steps, and performance metrics. Figure~\ref{fig:environment_visualization} shows visualizations of each environment.

\subsection{DeepMind Control Suite}
\label{appendix:environments_dmc}

The DeepMind Control Suite~\citep[DMC]{tassa2018dmc} is a standard benchmark for continuous control benchmarks with varying levels of complexity. Tasks in this benchmark range from simple low-dimensional ($\mathcal{S} \in \mathbb{R}^{3}$, $\mathcal{A} \in \mathbb{R}^{1}$) to highly complex continuous control ($\mathcal{S} \in \mathbb{R}^{223}$, $\mathcal{A} \in \mathbb{R}^{38}$). We evaluate 18 tasks, grouped into DMC Medium and DMC Hard. DMC Easy tasks are excluded due to their low difficulty. Full task lists appear in Tables~\ref{tab:appendix_dmc_medium_tasks} and~\ref{tab:appendix_dmc_hard_tasks}.

\subsection{Adroit}
\label{appendix:environments_adroit}

Adroit~\citep{rajeswaran2017adroit} comprises dexterous manipulation tasks involving in-hand manipulation, tool use, and articulated object control, performed using a 24 degree-of-freedom (DoF) anthropomorphic Shadow Hand. Due to the tasks' complexity, the state-of-the-art method DrM~\citep{xu2023drm} uses a privileged robot sensor vector alongside image observations (see official code: \url{https://github.com/XuGW-Kevin/DrM}). In contrast, we do not use privileged information. To increase difficulty, we also benchmark under sparse reward settings. The full task list is provided in Table~\ref{tab:appendix_adroit_tasks}.

\subsection{Meta-World}
\label{appendix:environments_metaworld}

Meta-World~\citep{yu2020metaworld} consists of 50 diverse robotic manipulation tasks using a simulated 7-DoF Sawyer arm in a tabletop setting. Following \citep{xu2023drm}, we select 8 tasks spanning object interaction, tool use, and precise motion control to cover a range of manipulation challenges. For the easier half, we replace the ground-truth dense reward function with a binary task completion signal (i.e. a \textit{sparse task completion reward}) and mark them as \texttt{sparse}. For details on the success metric, we refer the reader to~\citep{yu2020metaworld}. Full task list is provided in Table~\ref{tab:appendix_metaworld_tasks}.

\begin{table}[ht!]
\centering
\caption{\textbf{Environment details.} We list the episode length, frame stack, action repeat for each domain, total environment steps, and performance metrics used for benchmarking.}
\vspace{0.05in}
\begin{tabular}{lccc}
\hline
\noalign{\vspace{1.5pt}}
& \textbf{DMC} & \textbf{Adroit} & \textbf{Meta-World} \\
\noalign{\vspace{1.5pt}}
\hline
\noalign{\vspace{1.5pt}}
Frame stack      & $3$ & $3$ & $3$ \\
Action repeat       & $2$ & $2$ & $2$ \\
Episode length      & $1,000$ & $100$-$200$ & $500$ \\
Total env. steps    & $1$M & $1$M & $1$M  \\
Performance metric  & Average Return & Average Success & Average Success \\ \noalign{\vspace{1.5pt}}
\hline
\end{tabular}%

\label{tab:appendix_environment_details}
\vspace{-0.1in}
\end{table}

\clearpage

\begin{table}[ht!]
\centering
\parbox{0.9\textwidth}{
\caption{\textbf{DMC Medium Task List.} We evaluate 11 tasks from the DMC Medium benchmark, listed below. Performance for each task is reported at 1M environment steps.}
\label{tab:appendix_dmc_medium_tasks}
\centering
\vspace{0.05in}
\begin{tabular}{lcc}
\hline
\noalign{\vspace{1.5pt}}
\textbf{Task} & \textbf{State dim} $|\mathcal{S}|$ & \textbf{Action dim} $|\mathcal{A}|$ \\ 
\noalign{\vspace{1.5pt}}
\hline
\noalign{\vspace{1.5pt}}
\texttt{acrobot-swingup} & $6$ & $1$ \\
\texttt{cartpole-swingup-sparse} & $5$ & $1$ \\
\texttt{cheetah-run} & $17$ & $6$ \\
\texttt{finger-turn-easy} & $12$ & $2$ \\
\texttt{finger-turn-hard} & $12$ & $2$ \\
\texttt{hopper-hop} & $15$ & $4$ \\
\texttt{quadruped-run} & $78$ & $12$ \\
\texttt{quadruped-walk} & $78$ & $12$ \\
\texttt{reacher-easy} & $6$ & $2$ \\
\texttt{reacher-hard} & $6$ & $2$ \\
\texttt{walker-run} & $24$ & $6$ \\ 
\hline
\noalign{\vspace{1.5pt}}
\end{tabular}}
\end{table}

\begin{table}[ht!]
\centering
\parbox{0.9\textwidth}{
\caption{\textbf{DMC Hard Task List.} We evaluate 7 tasks from DMC Hard benchmark, listed below. Performance for each task is reported at 1M environment steps.}
\label{tab:appendix_dmc_hard_tasks}
\centering
\vspace{0.05in}
\begin{tabular}{lcc}
\hline
\noalign{\vspace{1.5pt}}
\textbf{Task} & \textbf{State dim} $|\mathcal{S}|$ & \textbf{Action dim} $|\mathcal{A}|$ \\ 
\noalign{\vspace{1.5pt}}
\hline
\noalign{\vspace{1.5pt}}
\texttt{dog-run} & $223$ & $38$ \\
\texttt{dog-trot} & $223$ & $38$ \\
\texttt{dog-stand} & $223$ & $38$ \\
\texttt{dog-walk} & $223$ & $38$ \\
\texttt{humanoid-run} & $67$ & $24$ \\
\texttt{humanoid-stand} & $67$ & $24$ \\
\texttt{humanoid-walk} & $67$ & $24$ \\ 
\noalign{\vspace{1.5pt}}
\hline
\end{tabular}%
}
\vspace{-0.1in}
\end{table}

\begin{table}[ht!]
\centering
\parbox{0.9\textwidth}{
\caption{\textbf{Adroit Task List.} We evaluate 3 tasks from Adroit benchmark, listed below. Performance for each task is reported at 1M environment steps.}
\label{tab:appendix_adroit_tasks}
\centering
\vspace{0.05in}
\begin{tabular}{lcc}
\hline
\noalign{\vspace{1.5pt}}
\textbf{Task} & \textbf{State dim} $|\mathcal{S}|$ & \textbf{Action dim} $|\mathcal{A}|$ \\
\noalign{\vspace{1.5pt}}
\hline
\noalign{\vspace{1.5pt}}
\texttt{door-v0-sparse} & $39$ & $28$ \\
\texttt{hammer-v0-sparse} & $46$ & $26$ \\
\texttt{pen-v0-sparse} & $45$ & $24$ \\
\noalign{\vspace{1.5pt}}
\hline
\end{tabular}}
\end{table}

\begin{table}[ht!]
\centering
\parbox{0.9\textwidth}{
\caption{\textbf{Meta-World Task List.} We evaluate 8 tasks from Meta-World benchmark, listed below. Performance for each task is reported at 1M environment steps.}
\label{tab:appendix_metaworld_tasks}
\centering
\vspace{0.05in}
\begin{tabular}{lcc}
\hline
\noalign{\vspace{1.5pt}}
\textbf{Task} & \textbf{State dim} $|\mathcal{S}|$ & \textbf{Action dim} $|\mathcal{A}|$ \\ 
\noalign{\vspace{1.5pt}}
\hline
\noalign{\vspace{1.5pt}}
\texttt{assembly} & $39$ & $4$ \\
\texttt{disassemble} & $39$ & $4$ \\
\texttt{pick-place-wall} & $39$ & $4$ \\
\texttt{stick-pull} & $39$ & $4$ \\
\texttt{coffee-push-sparse} & $39$ & $4$ \\
\texttt{hammer-sparse} & $39$ & $4$ \\
\texttt{soccer-sparse} & $39$ & $4$ \\
\texttt{sweep-into-sparse} & $39$ & $4$ \\ 
\noalign{\vspace{1.5pt}}
\hline
\end{tabular}%
}
\vspace{-0.1in}
\end{table}

\end{document}